\documentclass[a4paper,12pt]{article}
\usepackage[utf8]{inputenc}
\usepackage[english]{babel}
\usepackage{authblk}
\usepackage{graphicx}
\usepackage{mathptmx}
\usepackage[singlespacing]{setspace}
\usepackage[headheight=1in,margin=1in]{geometry}
\usepackage{fancyhdr}
\usepackage{hyperref} 
\usepackage{etoolbox}
\usepackage{subcaption}
\usepackage{url}

\usepackage{booktabs,dcolumn,caption}

\DeclareGraphicsExtensions{.jpg}

\graphicspath{{images/}}

\title{\vspace{-1cm}Deep Ensemble Learning for News Stance Detection}

\author[1]{Wenjun Liao}
\author[2]{Chenghua Lin}
\affil[1]{The Jheronimus Academy of Data Science, \href{mailto:w.liao@tue.nl}{\texttt{w.liao@tue.nl}}}
\affil[2]{University of Sheffield,  \href{mailto:c.lin@sheffield.ac.uk}{\texttt{c.lin@sheffield.ac.uk}}}

\date{}

\begin{document}

\maketitle
\thispagestyle{fancy}

\vspace{-3.5em}
\begin{center}
\newline
\end{center}

\section*{Extended Abstract}
Detecting stance in news is important for news veracity assessment because it helps fact-checking by predicting a stance with respect to a central claim from different information sources. Initiated in 2017, the Fake News Challenge Stage One\footnote{\url{http://www.fakenewschallenge.org}} (FNC-1) proposed the task of detecting the stance of a news article body relative to a given headline, as a first step towards fake news detection. The body text may agree or disagree with the headline, discuss the same claim as the headline without taking a position or is unrelated to the headline. 

Several state-of-the-art algorithms \cite{hanselowski2018retrospective,riedel2017simple} have been implemented based on the training dataset provided by FNC-1. We conducted error analysis for the top three performing systems in FNC-1. Team1 \textit{`SOLAT in the SWEN'} from Talos Intelligence\footnote{\url{https://github.com/Cisco-Talos/fnc-1}} won the competition by using a 50/50 weighted average ensemble of convolutional neural network and gradient boosted decision trees. Team2, \textit {`Athene'} from TU Darmstadt achieved the second place by using hard-voting for results generated by five randomly initialized Multilayer Perceptron (MLP) structures, where each MLP is constructed with seven hidden layers \cite{hanselowski2018retrospective}. The two approaches use features of semantic analysis, bag of words as well as baseline features defined by FNC-1, which include word/ngram overlap features and indicator features for polarity and refutation. Team3, \textit{`UCL Machine Reading'} uses a simple end to end MLP model with a 10000-dimension Term Frequency (TF) vector (5000 extracted from headlines and 5000 from text body) and a one-dimension TF-IDF cosine similarity vector as input features \cite{riedel2017simple}. The MLP architecture has one hidden layer with 100 units, and it's output layer has four units corresponding to four possible classes. Rectified linear unit activation function is applied on the hidden-layer and Softmax is applied on the output layer. The loss function is the sum of $l_2$ regularization of MLP weights and cross entropy between outputs and true labels. The result is decided by the argmax function upon output layer. Several techniques are adopted to optimize the model training process such as mini-batch training and dropout. According to our error analysis, UCL's system is simple but tough-to-beat, therefore it is chosen as the new baseline.  

\noindent\textbf{Method.}~~~In this work, we developed five new models by extending the system of UCL. They can be divided into two categories. The first category encodes additional keyword features during model training, where the keywords are represented as indicator vectors and are concatenated to the baseline features. The keywords consist of manually selected refutation words based on error analysis. To make this selection process automatic, three algorithms are created based on the Mutual Information (MI) theory. The keywords generator based on MI customized class (MICC) gave the best performance. Figure 1(a) illustrates the work-flow of the MICC algorithm. The second category adopts article body-title similarity as part of the model training input, where  word2vec is introduced and two document similarity calculation algorithms are implemented: word2vec cosine similarity and Word Mover's Distance. 

\noindent\textbf{Results.}~~~Outputs generated from different aforementioned methods are combined following two rules,   \textit{concatenation} and \textit{summation}. Next, single models as well as ensemble of two or three randomly selected models go through 10-fold cross validation. The output layer becomes $4\cdot\textit{N}$-dimension when adopting concatenation rule, where \textit{N} is the number of models selected for ensemble. We considered the evaluation metric defined by FNC, where the correct classification of relatedness contributes 0.25 points and correctly classify related pairs as agree, disagree or discuss contributes 0.75 points. Experimental results show that ensemble of three neural network models trained from simple bag-of-words features gives the best performance. These three models are: the baseline MLP; a model from category one where manually selected keyword features are added; a model from category one where added keywords feature are selected by the MICC algorithm. 

After hyperparameters tuning on validation set, the ensemble of three selected models has shown great performance on the test dataset. As shown in Table 1, our system beats the FNC-1 winner team Talos by 34.25 marks, which is remarkable considering our system's relatively simple architecture. Figure 1(b) demonstrates the performance of our system. Our deep ensemble model does not outstand in any of the four stance detection categories. However, it reflects the averaging outcome of the best results from the three individual models. It is the ensemble effect that brings the best result in the end. Evaluation has demonstrated that our proposed ensemble-based system can outperform the state-of-the-art algorithms in news stance detection task with a relatively simple implementation.

\begin{figure}
	\captionsetup{belowskip=-9pt}
	\begin{subfigure}{.45\textwidth }
		\centering
		\includegraphics[width=0.95\linewidth]{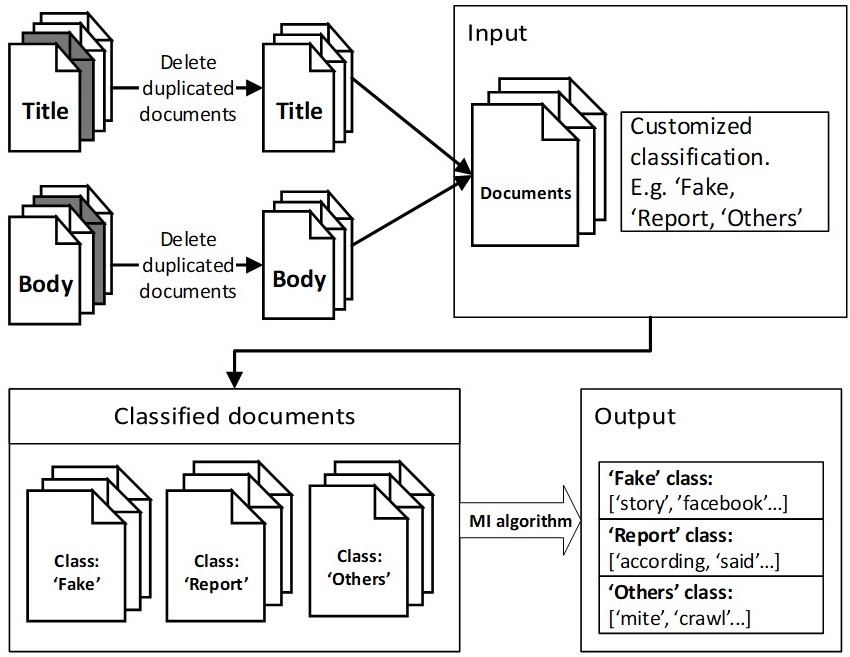}
		\caption{\centering}
		\label{fig:sfig1}
	\end{subfigure}%
	\begin{subfigure}{.55\textwidth}
		\centering
		\includegraphics[width=0.95\linewidth]{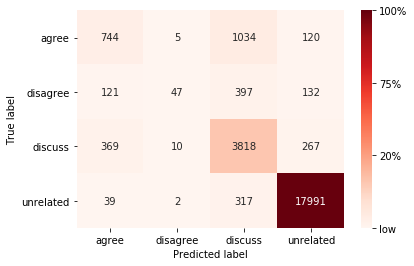}
		\caption{\centering}
		\label{fig:sfig2}
	\end{subfigure}

	\caption{(a) The illustration of customized-class based MI algorithm. The input is the customized theme word,  documents are then classified according to the themes. The output are groups of keywords under different class. (b) The heat map of the detection results. }
	\label{fig:fig}
\end{figure}

\begin{table}
	
	\newcolumntype{d}[1]{D{.}{.}{#1}} 
	\setlength\tabcolsep{0pt} 
	\caption{Performance comparison on the FNC-1 competition dataset}
	\label{turns}
	\begin{tabular*}{\textwidth}{@{\extracolsep{\fill}} l *{7}{d{2.3}} }
		\toprule
		
		Team & \multicolumn{4}{c}{Accuracy (\%)} & \multicolumn{1}{c}{\(F_1m\)} & \multicolumn{1}{c}{Relative Grade (\%)} & \multicolumn{1}{c}{Grade}  \\
		& \multicolumn{1}{c}{Agree} & \multicolumn{1}{c}{Disagree} & \multicolumn{1}{c}{Discuss}	& \multicolumn{1}{c}{Unrelated}\\
		
		\toprule
		UCL   & 0.44 & 0.066 & 0.814 & 0.979 & 0.404 & \multicolumn{1}{c}{81.72} & \multicolumn{1}{c}{9521.5} \\

		Athene  & 0.447 & 0.095 & 0.809 & 0.992 &0.416 & \multicolumn{1}{c}{81.97} & \multicolumn{1}{c}{9550.75} \\
				
		Talos 
		& 0.585 & 0.019 & 0.762 &0.987 &0.409& \multicolumn{1}{c}{82.02} & \multicolumn{1}{c}{9556.5} \\
				
		This work 
		& 0.391 & 0.067 & 0.855 &0.980 &0.403& \multicolumn{1}{c}{\textbf{82.32}} & \multicolumn{1}{c}{\textbf{9590.75}} \\
		
		\bottomrule
		
	\end{tabular*}
\vspace{-12pt}
\end{table}

\newpage
\lhead{}  
\rhead{}
	\bibliographystyle{plain}
	\bibliography{sample.bib}

\begin{thebibliography}{1}

\bibitem{hanselowski2018retrospective}
Andreas Hanselowski, Avinesh PVS, Benjamin Schiller, Felix Caspelherr, Debanjan
  Chaudhuri, Christian~M Meyer, and Iryna Gurevych.
\newblock A retrospective analysis of the fake news challenge stance detection
  task.
\newblock {\em arXiv preprint arXiv:1806.05180}, 2018.

\bibitem{riedel2017simple}
Benjamin Riedel, Isabelle Augenstein, Georgios~P Spithourakis, and Sebastian
  Riedel.
\newblock A simple but tough-to-beat baseline for the fake news challenge
  stance detection task.
\newblock {\em arXiv preprint arXiv:1707.03264}, 2017.

\end{thebibliography}
\newpage

\section*{Appendix\footnote{This page is not included in the submitted camera-ready version.}}

The code of this work is available at Github: \url{https://github.com/amazingclaude/Fake_News_Stance_Detection}. \\
\\
The full thesis regarding this work is available at ResearchGate: \url{https://www.researchgate.net/publication/327634447_Stance_Detection_in_Fake_News_An_Approach_based_on_Deep_Ensemble_Learning}.

\end{document}